\documentclass{article}

    \PassOptionsToPackage{numbers, compress}{natbib}

    \usepackage[preprint]{neurips_2019}

\usepackage[utf8]{inputenc} 
\usepackage[T1]{fontenc}    
\usepackage{hyperref}       
\usepackage{url}            
\usepackage{booktabs}       
\usepackage{amsfonts}       
\usepackage{nicefrac}       
\usepackage{microtype}      
\usepackage{graphicx}
\usepackage{amsmath}
\usepackage{booktabs}
\usepackage{algorithm}
\usepackage{algorithmic}
\usepackage{amsmath}

\title{InfoRL: Interpretable Reinforcement Learning using Information Maximization}

\author{
  Aadil Hayat \\
  IIT Kanpur\\
  \texttt{aadilh@iitk.ac.in} \\
   \And
  Utsav Singh \\
  IIT Kanpur \\
  \texttt{utsavz@iitk.ac.in } \\
  \And
  Vinay P. Namboodiri \\
  IIT Kanpur \\
  \texttt{vinaypn@iitk.ac.in} \\
}

\begin{document}

\maketitle

\begin{abstract}
  Recent advances in reinforcement learning have proved that given an environment we can learn to perform a task in that environment if we have access to some form of reward function (dense, sparse or derived from IRL). But most of the algorithms focus on learning a single best policy to perform a given set of tasks. In this paper we focus on an algorithm that learns to not just perform a task but different ways to perform the same task. As we know when the environment is complex enough there always exists multiple ways to perform a task. We show that using the concept of information maximization it is possible to learn latent codes for discovering multiple ways to perform any given task in an environment.
\end{abstract}

\section{Introduction}

Reinforcement learning has been able to achieve some really impressive results in the recent past. Beating humans at the game of Go \cite{44806}, performing complex robotic tasks \cite{levine2016end} and outperforming humans in Atari games \cite{mnih2013playing} are just some of the many achievements. The common approach of many reinforcement learning algorithms is to define a reward function and try to learn a policy for performing well on the task at hand by maximizing the reward function. If there are multiple ways of doing a task, the standard reinforcement learning approaches effectively learn an optimal way of accomplishing the task. But what if there are multiple near optimal ways of solving a task and we want to discover more than one such way of solving the task.
\par To this end, we devise an approach that enables us to learn multiple near optimal ways of accomplishing the task at hand. We call our approach: InfoRL. InfoRL uses latent code prediction for disentangling multiple near optimal policies for solving a given problem. The trick is to maximize the mutual information between a sampled latent code and the policy output, which enables the latent code to correspond to a particular policy, and hence provides a particular way of solving the problem at hand. The latent code effectively carries intrinsic information of solving the task in a particular way. Using this approach, all we need to do is sample a latent code, and we can output a policy which corresponds to a particular way of solving the task. This information maximization is done for all trajectories, which is computationally expensive. For solving this, we introduce an encoder-decoder like network for predicting latent codes which we use for predicting latent code, using the state and predicted action. This is explained in detail later.
\par The information maximization approach has provided significant results, when trying to learn disentangled representations output from a generative model in the InfoGAN \cite{chen2016infogan}. We choose to train a PPO policy \cite{schulman2017proximal} using the state and latent code representation, and output a reward function, which acts as the reinforcement signal to train the policy. The architecture is shown in Figure 1.
\par A salient feature of this approach is that we do not need any kind of supervision. We can set the problem in an unsupervised setting and are thus free to use any standard reinforcement learning algorithm of choice for training the policy, and we get a general continuous set of policies that are distinct from each other and are able to ultimately accomplish the task. 
\begin{figure}[t]
\includegraphics[scale=0.5]{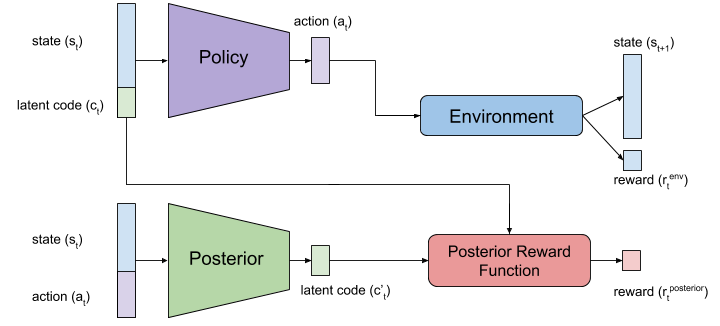}
\centering
\caption{InfoRL Architecture. The policy input is the concatenated state and latent code. The state and predicted action is input to the posterior network to predict the originally sampled latent code. The reward for the policy network is the combination of environment reward and reward for posterior network correctly predicting latent code.}
\label{figure:inforl-architecture}
\vspace{-5pt}
\end{figure}

\par We first discuss the general problem framework and discuss the relevant work done in this area. Then, we discuss the background required for understanding the approach used in the paper which is followed by a discussion of InfoRL algorithm in detail. In order to show that this approach does yield different policies, we perform extensive experimentation on a number of basic environments in simulation. We show that the environment agents do learn different ways of performing the task, and compare the performance on a number of performance metrics. 


\subsection{Relevant work}
Reinforcement Learning(RL) is the branch of machine learning where reward functions are used to generate supervised signals for training an agent towards solving a particular task. Some of the standard RL algorithms include the Q Learning \cite{watkins1992q} and policy gradient algorithm \cite{sutton2000policy}. RL algorithms combined with deep neural networks have been able to successfully solve a number of complicated tasks. These impressive results stem from the fact that neural networks act as excellent function approximators. Some such algorithms are DQN \cite{mnih2013playing}, DDPG \cite{lillicrap2015continuous}, proximal policy optimization(PPO) \cite{schulman2017proximal}, etc. PPO extends the idea of trust region update by using a surrogate training objective to improve policy gradient algorithm, and stabilizes its training by ensuring that the new policy learnt is close to the previous policy.
\par In this paper, we use the information maximization principle for maximizing the Shannon mutual information between the sampled posterior and the policy output to yield multiple policies for performing the task. It has been used with the standard RL setup for accomplishing a number of tasks. VIME \cite{houthooft2016curiosity} uses information maximization for generating an efficient exploration strategy based on maximization of information gain about the agent's belief of environment dynamics. In \cite{osa2019hierarchical}, the agent learns a latent variable using mutual information maximization to learn a hierarchical policy for solving the task at hand. Diversity is All You Need \cite{eysenbach2018diversity} uses information maximization to learn a maximum entropy policy, thus enabling the algorithm to explore and learn various skills in the environment, without supervision. Information maximization is also used in InfoGAN \cite{chen2016infogan} for learning disentangled representation from a given distribution, using generative adversarial networks \cite{goodfellow2014generative} for training on the data distribution.
\subsection{Contribution}
\par The major contributions of InfoRL are as follows:
\begin{enumerate}
    \item It allows the agent to choose from among the equally optimal different ways, some of which may be more suited to the specific task at hand.
    \item It gives a better understanding of how to accomplish a particular task by going through the available options.
    \item It is a step towards better visualizing and understanding how a reinforcement learning algorithm actually works towards achieving a goal.
\end{enumerate}
\section{Background}
In this section we discuss the principles and approaches used to build our model. We first discuss the standard reinforcement learning setup which is followed in the paper. We then discuss the information maximization principle used for generating the latent code for InfoRL.
\subsection{Reinforcement Learning}
We use the standard reinforcement learning setup in order to create a learning algorithm for solving a given problem. In this setup, the agent takes an action $a_t$ in either a fully observed state $s_t$ or a partially observed observation $o_t$, gets a reward $r_t$ and the environment changes its state to $s_{t+1}$. The return from the state is defined as the sum of the discounted future rewards, computed over a horizon T, i.e. $R_t=\sum_{i=t}^{T}\gamma^{i-t}r_i$. The learning agent's policy that it uses to take decisions is denoted by $a_t=\pi(s_t)$. The environment is modeled as a Markov Decision Process where the goal is to maximize the total reward during the course of learning over the episodes:
$J=\mathbb{E}_{r_i,s_i\sim\\E,a_i\sim\\\pi}[R_0]$. 
\par In this paper, we choose the on policy PPO algorithm as our default reinforcement learning algorithm. PPO uses trust region update to improve the policy using gradient descent, thus ensuring that the newly learnt policy is not radically different from the previous policy. This stabilizes the policy gradient algorithm and yields an efficient policy to solve the given task in continuous space. We add our model architecture as shown in Figure 1 on top of the PPO algorithm to enable learning multiple policies towards solving the task at hand. Note that we use an on policy algorithm for our experiments, but the algorithm can be easily extended to the off policy setting. This will be a part of our future work.
\subsection{Information Maximization}
Information maximization is the technique for maximizing the average mutual information between two function predictors. In this paper, we use it to maximize the information between the posterior function and the policy output (Figure 1). Using this, we aim to create a learning setting where the latent code corresponds to a particular learnt policy. 
\par Information entropy is the average rate at which information is produced from a data distribution. It is defined as:
$$H= -\sum_{i}P_i \log{P_i}$$
Mutual information between $A$ and $B$ can be shown as:
$$I(A|B)=H(A)-H(A|B)$$
InfoRL works by maximizing the mutual information between the posterior and the output policy.


\section{InfoRL}

When we have a complex enough environment for any reinforcement learning task there exist multiple ways to perform that task. This is because for a complex environment there might exist multiple trajectories which are near-optimal in terms of the reward function for the given task. Many of the current state-of-the-art reinforcement learning algorithms \cite{mnih2013playing}, \cite{44806}, \cite{levine2016end} explore the state and actions spaces of the environment but ultimately learn a policy function that produces a single near-optimal trajectory. Learning to produce multiple of these near-optimal trajectories is challenging because when we are learning to perform any RL task we do not have direct access to variability in the environment. So we need to proceed in an unsupervised manner to discover and disentangle these near-optimal trajectories.

In this section we propose an algorithm that discovers different latent factors that are responsible for the variation in these near-optimal trajectories and learns policies that can produce trajectories corresponding to these latent factors. We define the generative process for the near-optimal trajectory $\tau_{task}$ for a given task as: $s_0 \sim \rho_0$, $c \sim p(c)$, $\pi \sim p(\pi|c)$, $a_t \sim \pi(a_t|s_t)$, $s_{t+1} \sim P(s_{t+1}|a_t, s_t)$, $r^{env}_{t} \sim R_{environment}(s_t, a_t)$, where $\rho_0$ is the distribution of initial states, $p(c)$ is the prior distribution of latent codes, $p(\pi|c)$ is the distribution over near-optimal policies from which we can sample any policy $\pi$ corresponding to a given latent code $c$, $P(s_{t+1}|a_t, s_t)$ and $R_{environment}(s_t, a_t)$ are the state-transition model and reward function of the environment respectively. In these $p(\pi|c)$ is unknown which we need to learn and $p(c)$ we need to fix before we start the training.

\begin{algorithm}[t!]
\caption{InfoRL Algorithm}
\label{alg:inforl}
\textbf{Input}: Initial parameters of policy and posterior networks $\theta_0$ and $\phi_0$ respectively;\\\
\textbf{Output}: Learned policy $\pi_\theta$
\begin{algorithmic}[1] 
    \FOR{$i= 0,1,2,...$ }
        \STATE Sample a batch of latent codes: $c_i \sim p(c)$
        \STATE Sample trajectories $\tau_i \sim \pi_{\theta_i}(c_i)$
        \STATE Sample state-action pairs $\chi_i \sim \tau_i$
        \STATE Take a policy step from $\theta_i$ to $\theta_{i+1}$, using PPO update rule with the following reward for each $(s,a)$:$$ R_{environment}(s,a) + \lambda R_{posterior}(Q'_{\phi_{i}}(s,a),c_i)$$
        \vspace{-15pt}
        \STATE Update $\phi_i$ to $\phi_{i+1}$ to minimize the mean-squared error between $c_i$ and $Q'(s,a)$
    \ENDFOR
\STATE \textbf{return} $\pi_\theta$
\end{algorithmic}
\end{algorithm}

To solve the above mentioned problem we modify our policy function $\pi$ to have a dependence on latent code $c$ in addition to state $s$. But just introducing latent code $c$ in the policy function $\pi(a|s,c)$ is not enough as the function can just ignore $c$ and it would fail to represent the variation in trajectories. To make sure that policy function uses latent code $c$ as much as possible we use the concept of information-theoretic based regularization of the model which increases the mutual information between the latent code $c$ and the state-action pair $(s,a)$. This concept was introduced by InfoGAN \cite{chen2016infogan} where they used mutual information maximization between the latent code $c$ and generated output in order to learn to disentangle the latent factors in a given data distribution. This was later extended to GANs based imitation learning by InfoGAIL \cite{li2017infogail} where they learn to disentangle the latent factors present in the given expert trajectories. In particular this form of regularization maximizes $L_{I}(\pi, Q)$ which is variation lower bound of mutual information between latent codes and trajectories denoted as $I(c; \tau)$, where $Q(c|\tau)$ is an approximation of the true posterior $P(c|\tau)$. 
$$L_{I}(\pi,Q) = \mathbb{E}_{c \sim p(c), a\sim \pi(.|s,c)}[\log Q(c|\tau)] + H(c) \leq I(c;\tau)$$
But directly working with entire trajectories $\tau$ for the posterior approximation $Q$ is computationally expensive and it can only generate sparse rewards for episodes which is difficult to train. So instead we introduce a new function $Q'(s,a)$ similar to posterior $Q(c|s,a)$ which tries to predict the latent code $c$ given a state-action pair $(s,a)$. Using this predicted latent code and actual latent code for a given state-action pair $(s,a)$ we introduce a dense reward for each step of the episode. The policy function $\pi(s,c)$ and function $Q'(s,a)$ when trained jointly act as encoder-decoder pair where $\pi(s,c)$ encodes the information from state $s$ and latent code $c$ into action $a$ and $Q'(s,a)$ tries to reconstruct latent code $c$ given the state $s$ and generated action $a$. This architecture is shown in Figure \ref{figure:inforl-architecture}. Algorithm \ref{alg:inforl} represents a practical algorithm based on Proximal Policy Gradients which we call InfoRL. 

\section{Experiments}

\begin{figure*}[h!]
    \setlength{\fboxsep}{0pt}
    \setlength{\fboxrule}{0pt}
    \fbox{\includegraphics[trim={3cm 1cm 3cm 1cm},clip,scale=0.33]{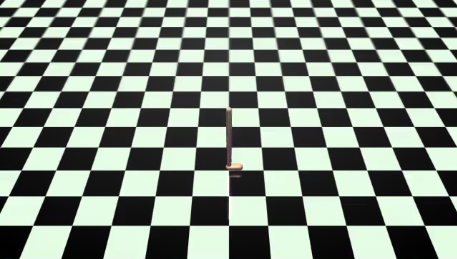}}   
    \hspace{8px}
    \fbox{\includegraphics[trim={2cm 0 2cm 0},clip,scale=0.25]{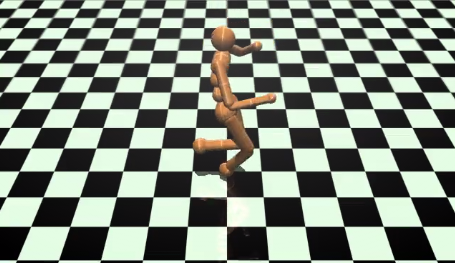}}   
    \hspace{8px}
    \fbox{\includegraphics[trim={2cm 0 2cm 0},clip,scale=0.25]{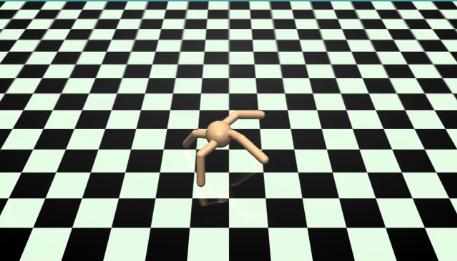}}
    \hspace{8px}
\fbox{\includegraphics[trim={2cm 6.5cm 2cm 2.5cm},clip,scale=0.14]{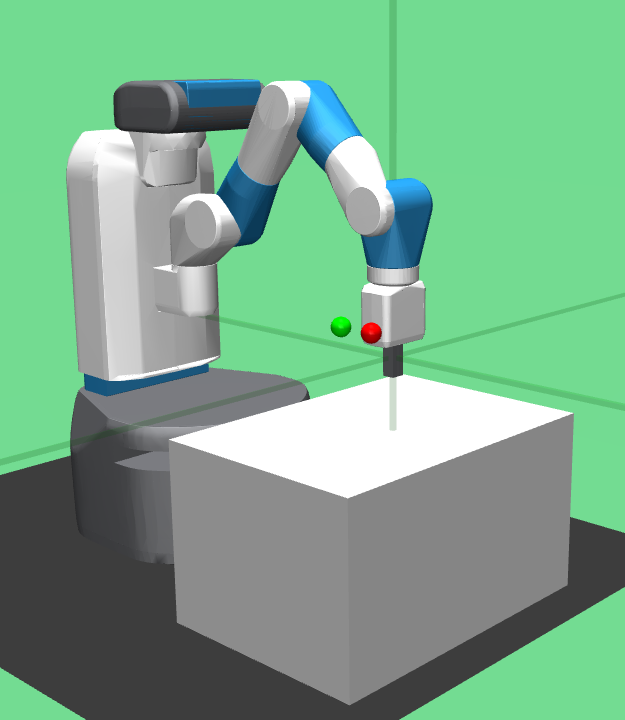}}
    \caption{\textbf{Sample environment images.} Left to right: \textit{Walker2dSpeed}, \textit{HumanoidSpeed}, \textit{AntDirection} and \textit{FetchTwoGoalReach}. }
    \label{figure:env-images}
\end{figure*}

To evaluate the performance of our method we created new benchmark environments by modifying existing OpenAI Gym environments. These new environments perform the same task as the original environments but include multiple equal reward trajectories to complete a particular task. The trajectories learned from InfoRL algorithm for these environments show that the model learns to discover and disentangle the variability present in these environments and learns to perform the given task in multiple ways controlled by the latent code.  The details of the environments are given below. Sample images from the environment can be found in Figure \ref{figure:env-images}. For all the following experiments we use a simple fully connected neural network with 2 hidden layers of 256 units each.

\subsection{Direction Experiment}

For the direction experiment we created a new environment named \textit{AntDirection-v1} which is based on the OpenAI's environment \textit{Ant-v2}. The \textit{Ant-v2} environment is a Mujoco based four-legged robotic environment where the reward function is designed so that the agent gets high reward for moving forward in the x-direction. We modify this reward function so that agent get high reward for moving forward in all direction of the 2-D plane. The idea behind this is to create a reward function that is agnostic to the direction in which the ant is moving and moving forward in any direction would result in equally rewarding trajectories. This can be seen from the following equations.

$$\text{Old forward reward at timestep } i = \dfrac{x_i - x_{i-1}}{dt}$$

\[\text{New forward reward at timestep }i =  \begin{cases} 
      \dfrac{||pos_i - pos_{i-1}||}{dt} & \text{if }||pos_i||>||pos_{i-1}||, \\
      0 & \text{otherwise.} 
   \end{cases}
\]
where $dt$ is the time duration between the steps of the environment, $x_j$ is the X-position of ant at timestep $j$, $y_j$ is the Y-position of ant at timestep $i$ and $pos_j = [x_j, y_j]$.

\begin{figure}[h!]
\includegraphics[trim={0.4cm 0 0 0},clip,scale=0.45]{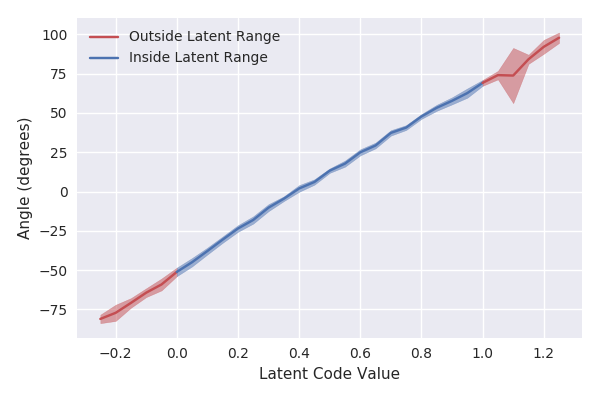}
\centering
\caption{\textbf{Ant direction experiments results.} We plot the variation of direction in which the ant moves with the continuous latent code in range $[-0.25, 1.25]$ at an interval of $0.05$. We run 10 episodes for each latent code. The direction is in terms of angle (in degrees) from the positive X-axis}
\label{figure:ant-direction}
\end{figure}

\begin{figure*}[h!]
    \setlength{\fboxsep}{0pt}
    \setlength{\fboxrule}{0pt}
    \fbox{\includegraphics[scale=0.27]{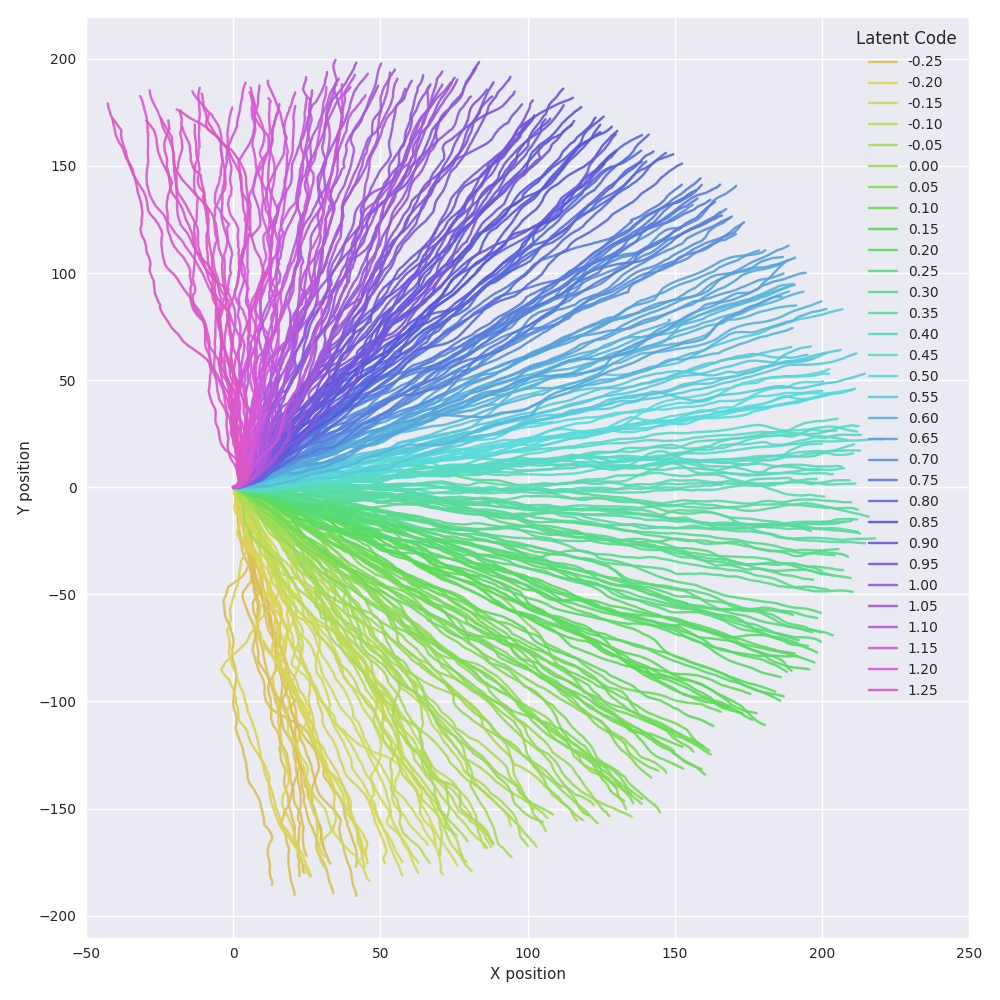}}   
    \fbox{\includegraphics[scale=0.27]{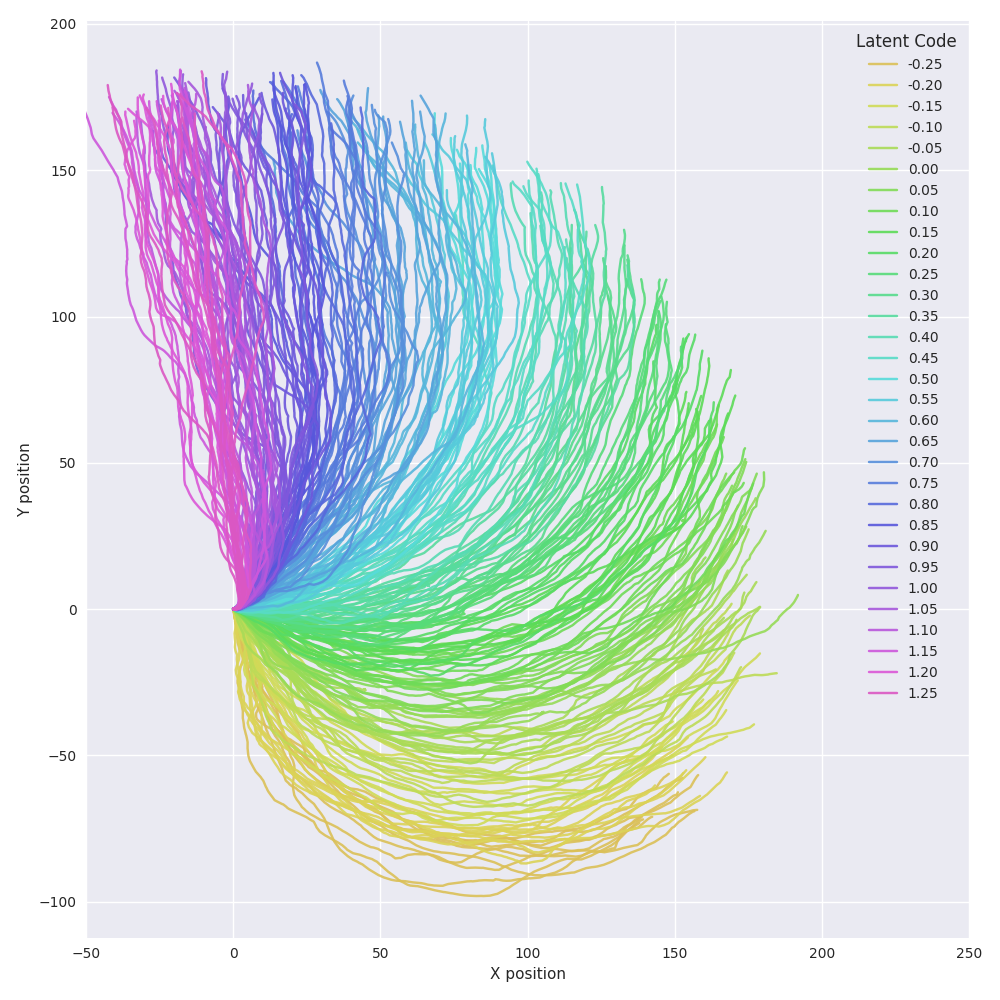}}
    \caption{\textbf{Ant direction experiment trajectories.} We plot the trajectories generated by the model trained using InfoRL on \textit{AntDirection-v1} environment on varying the continuous latent code. \textbf{Left:} Trajectories generated by fixing a specific latent code in range $[-0.25, 1.25]$ at an interval of $0.05$ for the whole episode. \textbf{Right:} Trajectories generated by fixing a starting latent code in range $[-0.25, 1.25]$ then increasing the latent code by $0.05$ upto $1.25$ after every $50$ steps.}
    \label{figure:ant-position-variation}
    \vspace{-5pt}
\end{figure*}

To discover and disentangle the direction of the trajectory we choose to model the latent distribution with one continuous code: $c \sim Uniform(0,1)$. After training the model on \textit{AntDirection-v1} environment with this latent code distribution using InfoRL algorithm \ref{alg:inforl} we see that the latent code captures variation in the direction of movement. We plot the variation with latent code of ant's movement direction in terms of angle (in degrees) from the positive x-axis shown in Figure \ref{figure:ant-direction}. We also plot the variation in the trajectories with variation in latent code in Figure \ref{figure:ant-position-variation}.

\subsection{Speed Experiments}

For the speed experiments we created new environments named \textit{Walker2dSpeed-v1} and \textit{HumanoidSpeed-v1} which is based on the OpenAI gym's environments \textit{Walker2d-v2} and \textit{Humanoid-v2} respectively. The \textit{Walker2d-v2} and \textit{Humanoid-v2} are Mujoco based robotic environment where the reward function is designed for the agent to move as fast as possible in the positive x direction. We modify these environments' reward function so that moving forward in x direction at different speeds gives same reward. This can be seen from the following equations.

$$\text{Old forward reward at timestep } i = \dfrac{x_i - x_{i-1}}{dt}$$

\[\text{New forward reward at timestep }i =  \begin{cases} 
      1 & \text{if }x_i-x_{i-1}>d_{threshold}, \\
      0 & \text{otherwise.} 
   \end{cases}
\]
where $dt$ is the time duration between the steps of the environment, $x_j$ is the X-position of robot at timestep $j$ and $d_{threshold}$ is the hyperparameter of the environment above which we consider that the robot is moving forward.

\begin{figure*}[h!]
    \setlength{\fboxsep}{0pt}
    \setlength{\fboxrule}{0pt}
    \fbox{\includegraphics[scale=0.45]{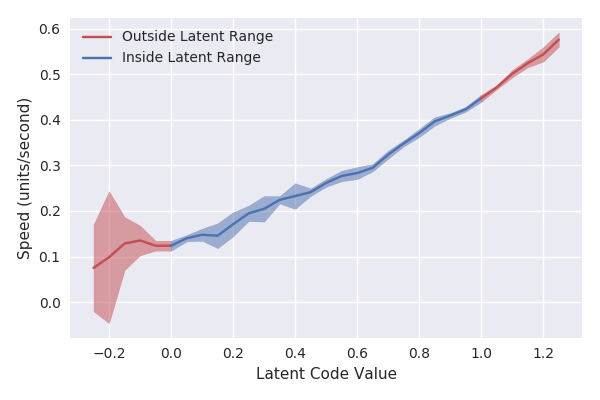}}   
    \fbox{\includegraphics[scale=0.45]{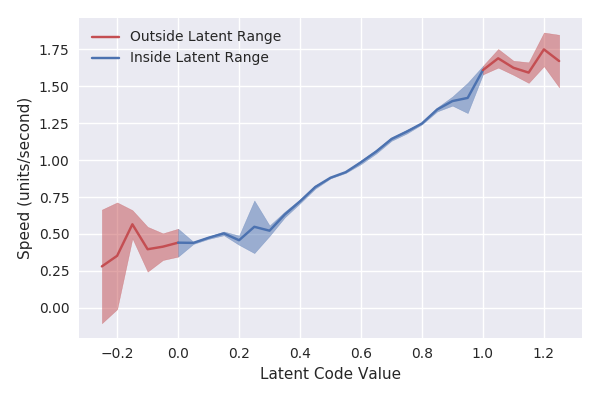}}
    \caption{\textbf{Speed experiment results.} We plot the variation of the speed with the continuous latent code in range $[-0.25, 1.25]$ at an interval of $0.05$. We run 10 episodes for each latent code. \textbf{Left:} \textit{HumanoidSpeed-v1} results. \textbf{Right:} \textit{Walker2dSpeed-v1} results. }
    \label{figure:speed-experiments}
    \vspace{-5pt}
\end{figure*}

To discover and disentangle the different speed of moving forward we choose to model the latent distribution with one continuous code: $c \sim Uniform(0,1)$. After training agents on the above mentioned environments with this latent code distribution we see that latent code captures variation in the speed of the agent. We plot the variation of speed with latent code as shown in Figure \ref{figure:speed-experiments}.



\subsection{Robotics Multi-Goal Experiments}

For the robotics experiment we created three new environments named \textit{FetchTwoGoalReachDense-v1}, \textit{FetchThreeGoalReachDense-v1} and \textit{FetchFourGoalReachDense-v1} which is based on the OpenAI gym's environment \textit{FetchReachDense-v1}. The \textit{FetchReachDense-v1} environment where we have a robotics agent and we have to move its end-effector to a randomly sampled goal position in 3D space. We modify this environment so that there are multiple goals which give equal rewards. 

\begin{figure*}[h!]
    \setlength{\fboxsep}{0pt}
    \setlength{\fboxrule}{0pt}
    \fbox{\includegraphics[scale=0.35]{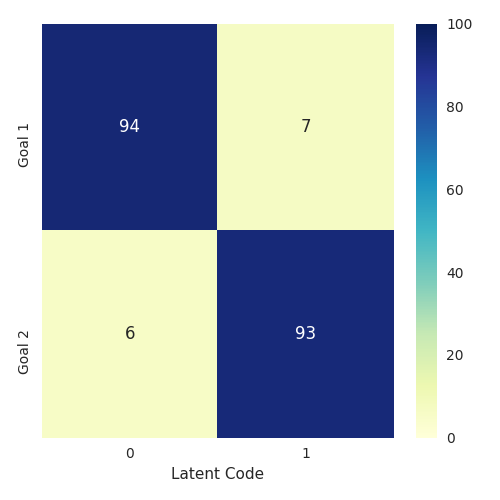}}   
    \hspace{4px}
    \fbox{\includegraphics[scale=0.35]{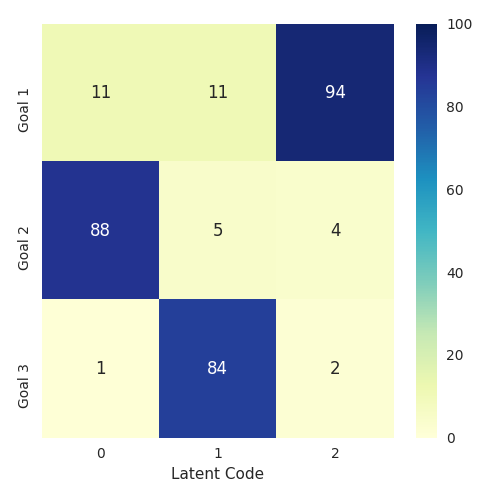}}
    \hspace{4px}
    \fbox{\includegraphics[scale=0.35]{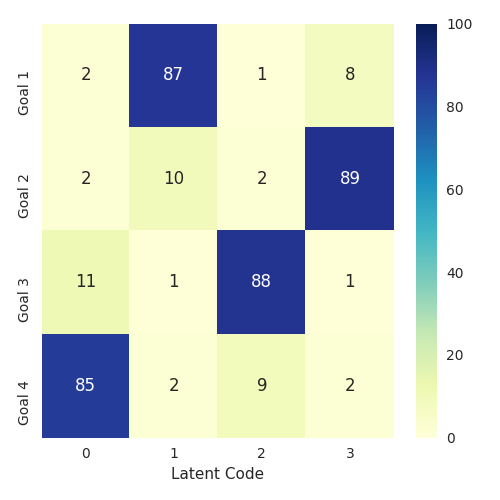}}
    \caption{\textbf{Experimental results for Robotics Multi-Goal Reach environments.} We plot the confusion matrix between the latent code for the episode and the goal near which the episode ends. We run the experiments for 100 episodes for each latent code. We set the number of categories for the latent code equal to number of goals in the environment. \textbf{Left:} \textit{FetchTwoGoalReachDense-v1} results. \textbf{Center:} \textit{FetchThreeGoalReachDense-v1} results. \textbf{Right:} \textit{FetchFourGoalReachDense-v1} results. }
    \label{figure:fetch-reach-experiments}
    \vspace{-5pt}
\end{figure*}

To discover and disentangle the trajectories for going to the different goals we choose to model the latent distribution with a categorical code: $c \sim Cat(K = N, p = 1/N)$ where $N$ is the number of goals. After training agent we plot the correlation between the latent code and goal near which the episode ends in Figure \ref{figure:fetch-reach-experiments}. We can see from the plot that the learned latent codes clearly disentangles reaching different goals.




\section{Conclusion}

This paper introduces a representation learning algorithm called InfoRL. This algorithm works with reinforcement learning tasks which have multiple ways to complete the task. It discovers and disentangles the multiple trajectories using which the task can be completed and it also learns a latent code to control over which trajectory to perform. We validate our approach over diverse set of reinforcement learning environments. This paper also contributes 6 different environments which can be used for benchmarking future work in this area of interpretable reinforcement learning.

\section{Future Work}

The proposed method presents a number of opportunities for further work.The experiments in this paper are done with an on-policy algorithm. The method can also be extended in an off-policy setting using any standard off policy RL algorithm. Since the latent code corresponds to an optimal policy, the method can be used in a hierarchical reinforcement learning setting. The method can also be applied to reduce the over-fitting problem in competitive self-play. We plan to explore some of the various application of InfoRL in different RL settings in the future.


\clearpage
\bibliographystyle{unsrtnat}
\bibliography{paper}

\end{document}